%% file: main.tex
\documentclass[letterpaper, 10 pt, conference]{ieeeconf}  

\IEEEoverridecommandlockouts                              

\usepackage{graphicx}
\usepackage{multirow}
\usepackage[dvipsnames]{xcolor}
\usepackage{color}
\usepackage{amsmath}
\usepackage{makecell}
\usepackage{booktabs}
\usepackage{wrapfig}
\usepackage{array}
\usepackage{hyperref}
\usepackage{cite}
\usepackage{pifont}
\usepackage{standalone}
\def\etal{\emph{et al.}}
\usepackage[dvipsnames]{xcolor}
\usepackage{fancyvrb}
\usepackage{makecell}
\DefineVerbatimEnvironment{HighlightVerbatim}{Verbatim}
  {commandchars=\\\{\},formatcom=\color{black}}
\hypersetup{
  colorlinks=true,
  linkcolor=red
}

\title{\LARGE \bf
QuadrupedGPT: Towards a Versatile Quadruped \\
Agent in Open-ended Worlds
}

\author{Yuting Mei$^{1\dag}$, Ye Wang$^{1\dag}$, Sipeng Zheng$^{2}$ and Qin Jin$^{1}$$^{*}$
\thanks{$^{1}$Renmin University of China}%
\thanks{$^{\dag}$Equal contributions}%
\thanks{$^{2}$Beijing Academy of Artificial Intelligence}%
\thanks{$^{*}$Corresponding author {\tt\small qjin@ruc.edu.cn}}%
}
\begin{document}

\maketitle
\thispagestyle{empty}
\pagestyle{empty}

\input{Sections/0_abstract}
\input{Sections/1_introduction}
\input{Sections/2_related_work}
\input{Sections/3_method}

\input{Sections/4_experiments}
\input{Sections/5_conclusion}

\bibliographystyle{IEEEtran}
\bibliography{reference}

\clearpage
\appendix
\input{Supplementary/Sections/all}
\end{document}

%% file: Sections/0_abstract.tex
\begin{abstract}

As robotic agents increasingly assist humans in reality, quadruped robots offer unique opportunities for interaction in complex scenarios due to their agile movement.
However, building agents that can autonomously navigate, adapt, and respond to versatile goals remains a significant challenge.
In this work, we introduce QuadrupedGPT designed to follow diverse commands with agility comparable to that of a pet. 
The primary challenges addressed include: i) effectively utilizing multimodal observations for informed decision-making; ii) achieving agile control by integrating locomotion and navigation; iii) developing advanced cognition to execute long-term objectives.
Our QuadrupedGPT interprets human commands and environmental contexts using a large multimodal model. Leveraging its extensive knowledge base, the agent autonomously assigns parameters for adaptive locomotion policies and devises safe yet efficient paths toward its goals. 
Additionally, it employs high-level reasoning to decompose long-term goals into a sequence of executable subgoals.
Through comprehensive experiments, our agent shows proficiency in handling diverse tasks and intricate instructions, representing a significant step toward the development of versatile quadruped agents for open-ended environments. 

\end{abstract}

%% file: Sections/1_introduction.tex
\section{Introduction}
\label{sec: intro}

Pets like cats and dogs are naturally agile creatures, yet they often require extensive training to understand and follow human commands. Inspired by this agility, researchers have made significant explorations in teaching quadruped robots basic skills such as locomotion and navigation~\cite{margolis2023walk,zhuang2023robot,yang2023iplanner,lee2020learning}.
However, most efforts fall short of developing a holistic agent capable of handling cognitively demanding tasks that require a seamless integration of these basic skills. 
As robotic agents become increasingly vital in dynamic real-world environments, the challenge lies not only in mastering movement and navigation, but also in enabling autonomous decision-making and long-term planning. 

Recent progress in large multi-modal models (LMMs)~\cite{gpt4o,zheng2024unicode,li2023blip} 
have highlighted their potential as versatile tools~\cite{liu2024visual,zhang2023video,roziere2023code,zheng2023steve,feng2023llama}, laying the foundation for the development of holistic agents~\cite{wang2023voyager}.
Quadruped robots, with their ability to emulate pet-like behaviors, offer a promising avenue for developing such agents  that can operate autonomously in the real world. 
Considering this, we ask: \textit{``Could we develop an LMM-enpowered agent that brings together the agility of pets with the cognitive abilities of humans, enabling it to interact with our open-ended world without human intervention?''}

To achieve this goal, several key challenges must be addressed:
\ding{182} The quadruped agent should be able to understand both human commands and its environment, while  dynamically processing multimodal observations.
\ding{183} To replicate agility of four-legged animals, the agent must flexibly adapt its locomotion strategies to navigate diverse and challenging terrains, while simultaneously identifying efficient and safe paths.
\ding{184} To handle complex, long-term goals, the quadruped agent must be equipped with advanced reasoning capabilities to perform high-level planning. 

\begin{figure}[t]
  \centering
  \includegraphics[width=\linewidth]{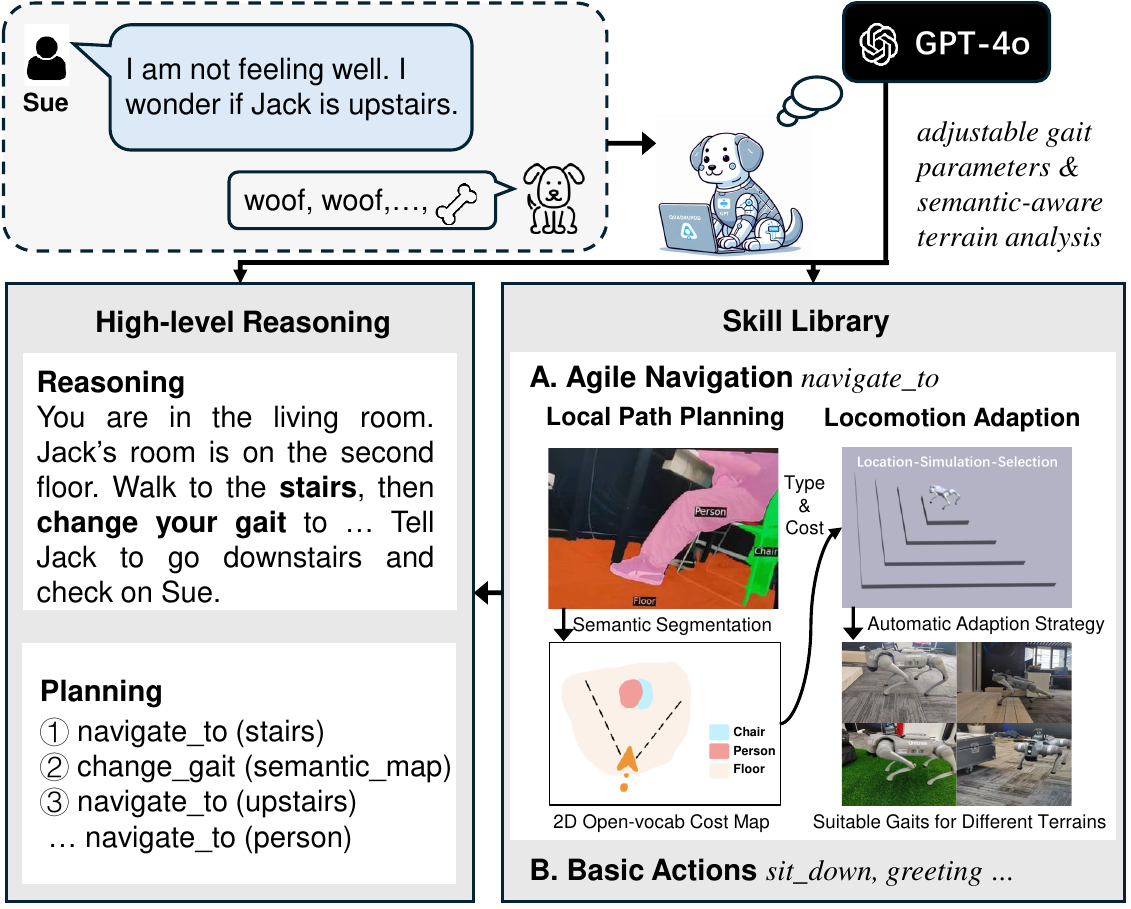}
  \caption{Built upon the cutting-edge large multimodal model, Quadruped-GPT aims to develop a versatile quadruped agent with the agility of four-legged pets while being able to comprehend intricate human commands and complete them safely and efficiently in open-world environments.}
  \label{fig:intro}
\end{figure}

To address these challenges, we propose \textbf{QuadrupedGPT}, a versatile quadruped agent designed to combine agility with advanced cognition, enabling it to tackle complex, goal-driven tasks across diverse scenarios. 
Specifically, \ding{182} The agent leverages powerful foundation models to reason about the environment and interpret commands.
\ding{183} To match the agility of pets, it utilizes the LMM to enhance locomotion and local navigation performance. 
We design a Location-Simulation-Selection (LSS) strategy that autonomously selects optimal gaits for varying terrains, eliminating the need for manual step-by-step adjustments.
Additionally, QuadrupedGPT maintains an open-vocabulary semantic cost map for path planning, allowing it to effectively assess terrain traversability and detect obstacles.
\ding{184} For long-horizon tasks, our agent decomposes complex objectives into a sequence of executable subgoals via LMM reasoning.

In summary, our contributions are as follows: 
i) We propose QuadrupedGPT, a versatile agent powered by cutting-edge foundation models to understand this world and perform complex tasks.
ii) Our agent achieves agile control similar to quadruped pets by leveraging LMMs for adaptive locomotion and intelligent path planning.
iii) Extensive experiments on benchmarks, including locomotion adaptation, local navigation, and high-level reasoning, demonstrate its capabilities as a general-purpose agent in the open-ended world.

%% file: Sections/2_related_work.tex
\section{Related Work}
\label{sec: related work}

\subsection{Quadruped locomotion}

Traditional approaches divide locomotion into multiple phases~\cite{gangapurwala2022rloc,fahmi2022vital}.
Since the design of such systems is tedious, end-to-end reinforcement learning (RL) approaches~\cite{margolis2023walk,kumar2021rma} train a policy to perform a direct mapping from proprioceptive information to joint control signals.
Later, some works explore incorporating visual input in locomotion policies~\cite{loquercio2023learning,kumar2021rma} such as ego-vision~\cite{agarwal2023legged,loquercio2023learning} and front-facing depth~\cite{cheng2023extreme,zhuang2023robot,hoeller2024anymal}.
Due to remarkable progress achieved by LMMs in diverse areas, recent works~\cite{shek2023lancar} explore using LMM to adapt locomotion strategies.
Shek~\etal~\cite{shek2023lancar} integrate an environment translator with RL agents, allowing environment-aware locomotion. However, their approach requires simultaneous training of the LMM and the policy, which is expensive.
Compared with traditional~\cite{margolis2023walk} and LMM-based RL approaches~\cite{shek2023lancar}, we design an automatic locomotion adaptation strategy that avoids tedious parameter tuning and policy training, offering a more efficient solution.

\subsection{Local path planning}

Previous approaches~\cite{paden2016survey,ratliff2009chomp,karaman2011sampling} follow a modularized paradigm by combining a traversability estimation module with a path-searching algorithm.
Instead, recent researches start to explore data-driven approaches, which can be broadly categorized into four types: 
1) modular geometric methods~\cite{cao2022autonomous} which rely primarily on geometric information~\cite{frey2022locomotion,hudson2021heterogeneous} to analyze environments and predict traversability based on specific metrics~\cite{fan2021step};
2) modular semantic methods~\cite{cai2022risk,maturana2018real}, which utilize additional semantic cues to assess a traversability cost for each semantic class; 
3) imitation and self-supervised learning~\cite{loquercio2021learning,pfeiffer2017perception} based on expert demonstrations;  and
4) RL-based methods. 
In addition to these approaches, Yang~\etal~\cite{yang2023iplanner} propose treating path planning as an offline bi-level optimization problem.
Roth~\etal~\cite{roth2023viplanner} further improves \cite{yang2023iplanner} by generating local paths leveraging geometric and semantic information.
Inspired by these works, we incorporate semantic information into local navigation to find a safe and efficient path.

\subsection{Large multimodal model for high-level reasoning}

Empowered by the reasoning capabilities of LMMs~\cite{brown2020language,raffel2020exploring,wei2022emergent}, research in embodied agents~\cite{duan2022survey,ravichandar2020recent,collins2021review} has increasingly utilized LMMs for high-level planning~\cite{huang2022language,min2021film}.
Given a task instruction, these works typically generate plans via strategies like skill decomposition~\cite{ouyang2024long,wang2023voyager,huang2022inner}. 
However, such plans can be impractical due to hallucinations~\cite{bang2023multitask} and not feasible. 
To address this, some works~\cite{liang2023code,huang2022inner,brohan2023can} propose robotic affordance prediction to decide the feasibility of an action~\cite{brohan2023can}.
In this paper, we leverage the advanced reasoning capability of LMMs to interpret and decompose human instructions. 
Combining this with automatic locomotion adaptation and semantic-aware path planning, our quadruped agent can achieve complex tasks.

%% file: Sections/3_method.tex
\section{Method}
\label{sec: method}

As shown in  Figure~\ref{fig:intro}, we present QuadrupedGPT, a general-purpose agent designed to autonomously solve complex, long-term tasks with the agility of natural four-legged animals. Unlike traditional quadruped robots, which are typically limited to basic locomotion and navigation, we integrate multimodal observations, adaptive control, and high-level cognition into a unified agent.
Our QuadrupedGPT includes three key components: i) in Section~\ref{sec:locomotion}, we introduce an automatic locomotion adaptation method that dynamically adjusts to different terrains without manual intervention. ii) in Section~\ref{sec:path_plan}, we detail the local path planning module, which allows QuadrupedGPT to navigate efficiently through diverse environments using a 2D open-vocabulary cost map. 
iii) in Section~\ref{sec:long_term}, we describe how the agent employs LMM-based high-level reasoning to break down long-term tasks into executable skill sequences, ensuring both agility and cognitive capability in achieving complex goals. 

\subsection{Automatic Locomotion Adaption}
\label{sec:locomotion}

A quadruped agent must navigate diverse and unseen terrains smoothly. 
One common approach is to train an all-encompassing locomotion policy using reinforcement learning. However, such RL-based policy often incur high training costs and offer limited generalization~\cite{cobbe2019quantifying}. 
Another approach, as proposed by~\cite{margolis2023walk}, is to incorporate multiple adjustable behavior parameters into the policy input, allowing the agent to adjust its actions dynamically. 
While~\cite{margolis2023walk} successfully trains a policy that encodes a structured family of locomotion strategies, it still requires manual adaption of behavior parameters, which limits full automation. 
To address this, we introduce  our automatic locomotion adaptation strategy. For better understanding, we first present a manual locomotion adaptation approach as a comparison.

\noindent\textbf{Manual locomotion adaption. } 
The locomotion policy $\pi(\cdot|c_t, b_t)$ is introduced in \cite{margolis2023walk}, where $c_t$ denotes the three-dimensional command vector $(v_x^{\rm cmd}, v_y^{\rm cmd}, w_z^{\rm cmd})$, controlling omnidirectional velocity, and $b_t$ represents adaptable behavior parameters that dictate how the robot moves. 
This policy requires manual intervention to set appropriate behavior parameters $b_t$ in real time, which is both labor-intensive (i.e., parameter tuning through trial and error) and inefficient (i.e., manual adaption is required for each new environment). 
To overcome this, we propose an automatic locomotion adaptation method by using LMM, significantly improving the system's efficiency and autonomy. 

\noindent\textbf{Adaptable parameters. } 
The adaptable behavior parameters $b_i$=$[h_z, f, \phi, s_y, h_z^f, \theta]$ include body height, stepping frequency, body pitch, foot stance width, foot swing height, and timing offsets between foot pairs, respectively. 
We refer further details about policy implementation to \cite{margolis2023walk}. 
These behavior parameters directly control the quadruped agent's locomotion style. We find that LMMs encode rich common sense knowledge of locomotion actions. For example, it can determine that a quadruped agent should lower its body height and raise its foot swing height when climbing stairs. However, LMMs are often not sensitive to precise numerical values, so we design our locomotion adaptation approach accordingly.

\begin{figure}
\centering
  \includegraphics[width=0.9\linewidth]{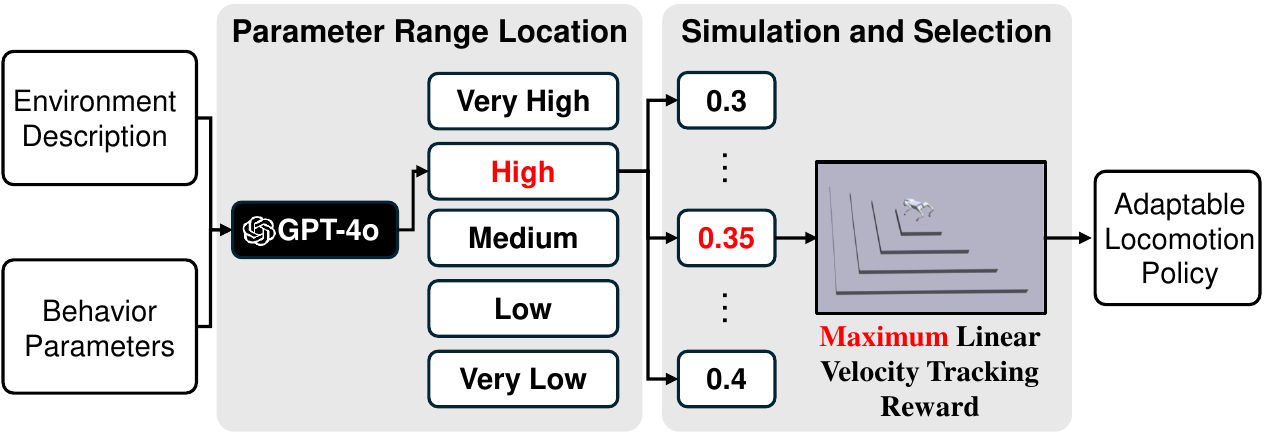} 
  \caption{The overview of our automatic locomotion adaption strategy ``Location-Simulation-Selection'' (LSS).}
\label{fig:method_locomotion} 
\end{figure}

\noindent\textbf{Automatic locomotion adaption based on LMM. }
To achieve automatic locomotion adaptation, we propose the ``Location-Simulation-Selection'' (LSS) strategy inspired by~\cite{shek2023lancar}. As illustrated in Figure~\ref{fig:method_locomotion}, LSS consists of three stages: 
i) \textit{Parameter Range Location}: Initially, the LMM estimates approximate parameter ranges (very low, low, medium, high, and very high) instead of exact values.  
ii) \textit{Simulation and Evaluation}: Candidate parameter sets are then sampled within the located ranges and evaluated in a simulation environment~\cite{makoviychuk2021isaac} based on cumulative rewards. 
iii) \textit{Best Parameter Selection}: Finally, the set of parameters yielding maximum linear velocity tracking reward is selected.

\begin{figure}
\centering
  \includegraphics[width=0.9\linewidth]{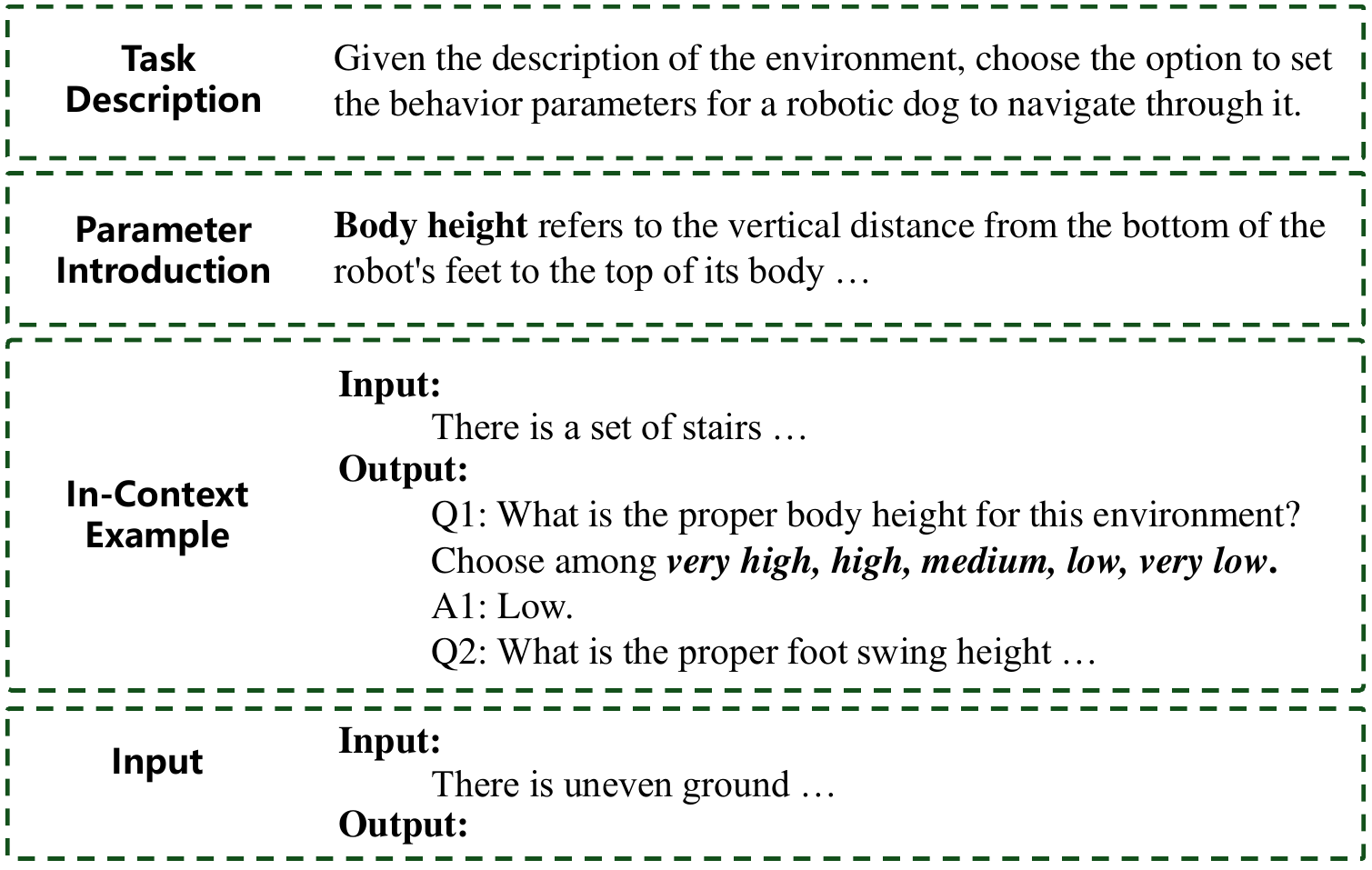} 
  \caption{Example prompt to guide quadruped agents across uneven terrains. 
  }
\label{fig:method_prompt} 
\end{figure}

To facilitate this process, 
as shown in Figure~\ref{fig:method_prompt}, we carefully design task-specific prompts to include a detailed task and behavior parameter description. 
We prompt the LMM to decide behavior parameters $b_t$ instead of depicting the environment~\cite{shek2023lancar}, as this more directly controls the robot's behavior.
Our strategy ensures QuadrupedGPT can autonomously adjust its behavior suitable for different terrains without manual intervention.

\noindent\textbf{Bridging simulation and reality. }
For real-world experiments, we additionally prompt the LMM to choose the most suitable simulation terrain according to the robot's environment and generate environment descriptions, ensuring a smooth transition from simulated to real environments.

\subsection{Local Path Planning}
\label{sec:path_plan}
Beyond locomotion, the robot must also be capable of planning efficient paths and adjust its gaits accordingly. 
Our navigation system is built upon~\cite{chang2023goat}, which divides navigation into global and local path planning.
Global path planning sets long-term target locations $(x,y)$, while local path planning focuses on identifying  trajectory waypoints to reach the target location $(x,y)$.
The complex and dynamic environment and diverse instruction requirements complicate this process, as
reaching the destination typically involves multiple trajectories, and different waypoints may require distinct gaits. The agent needs to identify the most suitable and collision-free next waypoint and the corresponding gait from its current location.
Existing methods, such as
\cite{chang2023goat},  cannot select paths based on human instructions and uses a fixed gait for different trajectory points.
Moreover, it depends on an instance segmentation module limited to predefined closed-set classes, restricting its adaptability to diverse and changing environments.
To overcome this, 
we propose using LMMs to enhance adaptability by combining dense segmentation and constructing cost maps for open-vocabulary objects and terrains. Additionally, we integrate LSS for selecting appropriate gaits during terrain transitions.

\noindent\textbf{Preliminaries. }
The environment where the robot operates is defined as 3-dim space $Q \in R^3$.
Within this space, $Q_{\text{obs}}$ denotes areas with non-traversable geometric or semantic obstacles, while $Q_{\text{trav}}$ encompasses the traversable regions where movement is safe. 
Inspired by prior works~\cite{yang2023iplanner,roth2023viplanner}, we construct a cost map that partitions the traversable area $Q_{\text{trav}}$ into $K$ subsets,
each with a navigable cost per pixel. These costs, scored from 0 to 1, reflect the difficulty or resource expense of traversing each subset, assisting the agent in selecting the most suitable path.

\begin{figure}[t]
\centering
  \includegraphics[width=0.9\linewidth]{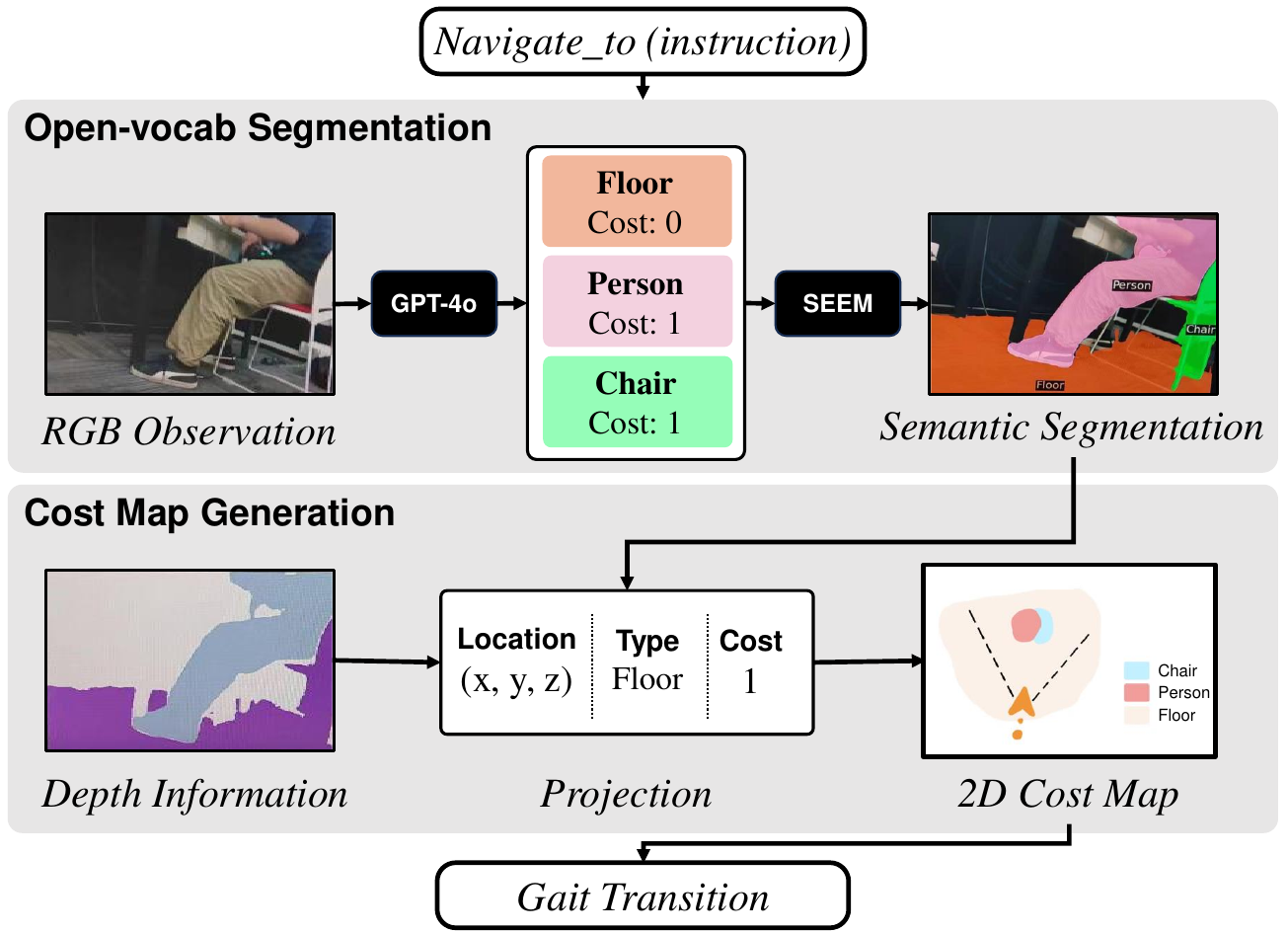} 
  \caption{Illustration of 2D open-vocabulary cost map generation.}
\label{fig:method_path} 
\end{figure}

\noindent\textbf{2D instance map generation. } 
We construct a 2D instance map as a spatial representation of the environment, which tracks the locations of object instances, obstacles, and explored areas.
Each cell on this map corresponds to an area of $25 cm^2 (5 \times 5cm)$ in the real world.
This map is generated by projecting egocentric semantic segmentation into point clouds using dense depth information captured by RGB-D cameras. 
These point clouds are then categorized into a 3D semantic voxel map. 
Summing along the vertical dimension produces the 2D instance map.

\noindent\textbf{2D open-vocabulary cost map. }
After generating the instance map, we leverage the LMM to generate an open-vocabulary cost map for the downstream path planner~\cite{sethian1996fast}. As shown in Figure ~\ref{fig:method_path}, the LMM first generates a comprehensive list of object and terrain categories based on the RGB observation of the environment.
Then, open-vocabulary dense segmentation~\cite{zou2023segment} is performed using these class names.
The LMM also assigns cost values to each object and terrain  based on common sense and user instructions. For example, for obstacles that are totally not travelable based on common sense, a cost of 1 ensures that these areas will not be considered during local path planning.
If the user specifically asks the agent to traverse certain terrains, the cost could be set to 0. If the user expects safer actions, the agent would tend to set the cost of slightly challenging terrains to 1.

Similar to 2D instance map generation, this process results in a 2D semantic cost map, which is instrumental in guiding our agent's navigation.
Equipped with this map, QuadrupedGPT can efficiently plan suitable trajectory waypoints towards its objectives while avoiding obstacles.

\noindent\textbf{Gait transition. }
The gait is determined in conjunction with the semantic cost map. When the robot enters a new terrain on the map, LSS is invoked to generate a suitable gait based on the current RGB observation.

\subsection{High-level Reasoning}
\label{sec:long_term}
The above modules: automated locomotion adaptation and local path planning, are crucial for endowing quadruped robots with agile moving on diverse terrains, while providing the required high-level cognitive capability: \textit{navigate\_to(instruction)}. 
This capability enables the robot to autonomously plan and execute corresponding motion strategies based on user instructions, such as  ``safely find the table'' or  ``reach the lawn by bypassing the stairs'', achieving flexible navigation. We further expand the quadruped robot's behavioral repertoire by integrating a series of basic actions, such as  \textit{sit\_down()}, \textit{lift\_up\_paw()}, \textit{shake\_body()} and \textit{speaking()}, into its skill library, allowing the robot to behave more like an intelligent pet. 
Upon receiving a new user instruction, the LMM parses and decomposes the task into a sequence of executable atomic skills, which the robot then executes and interacts with the environment in real-time through onboard sensors and actuators, ultimately achieving the goal specified by the user instruction.

%% file: Sections/4_experiments.tex
\section{Experiments}
\label{sec: exp}

\subsection{Experimental Details and Setup}
In this study, we choose Unitree Go2 as our platform, which is equipped with a RealSense D435i camera for RGB-D observation and an NVIDIA Orin-X for deploying the agent.
In addition, we adopt GPT-4o~\cite{gpt4o} as the LMM due to its powerful capacity.
For the locomotion adaptation benchmark, we conduct simulations in Isaac Gym~\cite{makoviychuk2021isaac}, averaging results over 10 runs, with each evaluation episode comprising 250 steps.
The benchmark tests the agent on five challenging terrains: uphill slopes, downhill slopes, upward stairs, downward stairs, and randomly generated uneven ground. 
We evaluate the agent based on the average reward for achieving commanded linear and angular velocities and maintaining the appropriate gait. 
We adopt the adjustable policy from~\cite{margolis2023walk}, setting the commanded linear velocity at $1.0\ m/s$ along the x-axis and the angular velocity at $0\ rad/s$ during evaluation, indicating that the robot should walk in a straight line. 
For real-world locomotion experiments, we compare our LSS with averaged-parameter gaits in two challenging scenarios: uphill slopes and upward stairs. We report the success rate for each setting over 5 runs. 
Lastly, we use SEEM ~\cite{zou2023segment} for open-vocabulary semantic segmentation.

Specifically, for the locomotion tasks in simulation, we utilize the rewards $\displaystyle r_{v_{x,y}^{\rm cmd}}$, $\displaystyle r_{\omega_z^{\rm cmd}}$, $\displaystyle r_{c_f^{\rm cmd}}$, $\displaystyle r_{c_v^{\rm cmd}}$ as evaluation metrics, as outlined in Table~\ref{tab:reward}. 
To implement manual parameter tuning (``Manual''), we recruit 10 participants and provide them with details and visual aids regarding behavior parameters and terrains. 
Each participant adjusts these parameters for specific tasks, with final results calculated by averaging the outcomes from 10 runs for each parameter set.
Additionally, an expert familiar with robot kinematics is also invited to perform the manual tuning (``Expert'').

\input{table/reward}

For real world locomotion experiments, we evaluate two scenarios: uphill slope and upward staircase. Success is defined as the robot reaching the platform beyond the slope or stairs. 
Each experiment is conducted 5 times, and the success rate is reported. We compare the adapted locomotion strategy with a fixed-parameter gait, where all behavior parameters are set to the average value within the adjustable range. 
Additionally, We test QuadrupedGPT's local path planning capabilities by navigating towards various target objects in a real world environment. A trial is considered successful if the agent reaches the target within a 0.5-meter radius.

\input{table/locomotion}

\subsection{Results of Locomotion Adaption}

\noindent\textit{\textbf{Manual tuning can be less effective than automatic approach without expert knowledge in practice. }}
As shown in Table ~\ref{tab:locomotion}, our automatic locomotion adaption strategy (``Auto''), which leverages the LMM to predict parameters values, outperforms manual tuning (``Manual'') in identifying optimal gait parameters.

\noindent\textit{\textbf{LSS strategy outperforms directly predicting the parameters. }}
The LMM's ability to handle numerical values related to the physical world is limited. However, as shown in the Table \ref{tab:locomotion}, after integrating the Location-Simulation-Selection strategy, our approach (``Auto+LSS'') surpasses even expert-tuned parameters (``Expert''). 
Thees comparison results suggest that with a carefully curated prompting process, the LMM can act as an autonomous specialist, even outperforming human experts.
We observe significant improvement in both $\displaystyle r_{v_{x,y}^{\rm cmd}}$ and $\displaystyle r_{\omega_z^{\rm cmd}}$, indicating our approach not only better achieves the velocity command but also enhances overall agility.

\begin{figure}[t]
  \centering
  \includegraphics[width=0.85\linewidth]{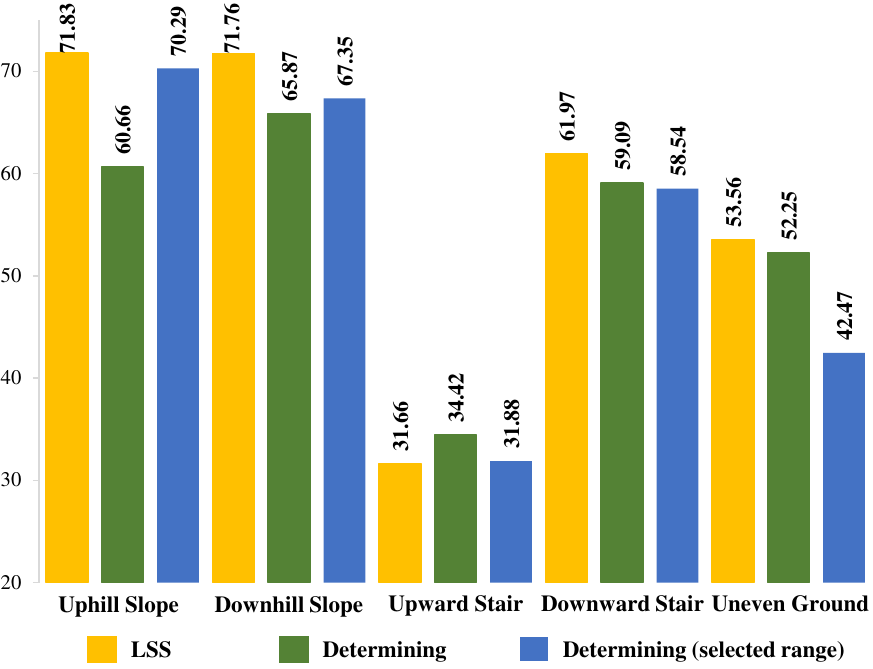}
  \caption{Ablation results of different parameter selection alternatives for LSS strategy. Average $r_{v_{x, y}^{cmd}}$ is reported.}
  \label{fig:ablation}
\end{figure}

\noindent\textbf{\textit{Sampling and simulating within the parameter range is crucial. }}
To investigate the impact of different approaches for selecting behavior parameters for the policy, we conduct ablation experiments as shown in Figure ~\ref{fig:ablation}.
In the LSS with the ``Sampling'' option, the LMM first locates a range, samples parameters within the range, and then simulates to determine the best parameter set. In contrast, 
LSS with the ``Determining'' option has the LMM choose directly among the midpoints of these ranges (i.e., options are given as numbers) without simulation. 
The LSS with the ``Determining" (selected range) option involves the LMM first locating a range, then selecting directly among the quintile points of these ranges.
The results demonstrate that simulation is crucial for validating parameters. 
Allowing the LMM to sample from ranges and simulate generally yields better performance compared to directly choosing values. 
Furthermore, LSS with the ``Determining" option shows notable instability, likely due to the LMM's limited sensitivity to numerical values. 
Therefore, we opt for the LSS with the sampling strategy, which consistently provides more stable and effective results.

\noindent\textbf{\textit{LSS outperforms fixed-parameter gait in real world experiments. }} 
As shown in Table \ref{tab:real_world_loco}, LSS achieves a higher success rate than the fixed-parameter gait in both challenging scenarios. Particularly, in the upward stairs scenario, the robot needs to adjust its leg height and body height for successful locomotion, which the fixed-parameter gait fails to accommodate. The failures observed with LSS were primarily due to some excessively large movements of the policy.

\begin{figure}[t]
  \centering
  \includegraphics[width=\linewidth]{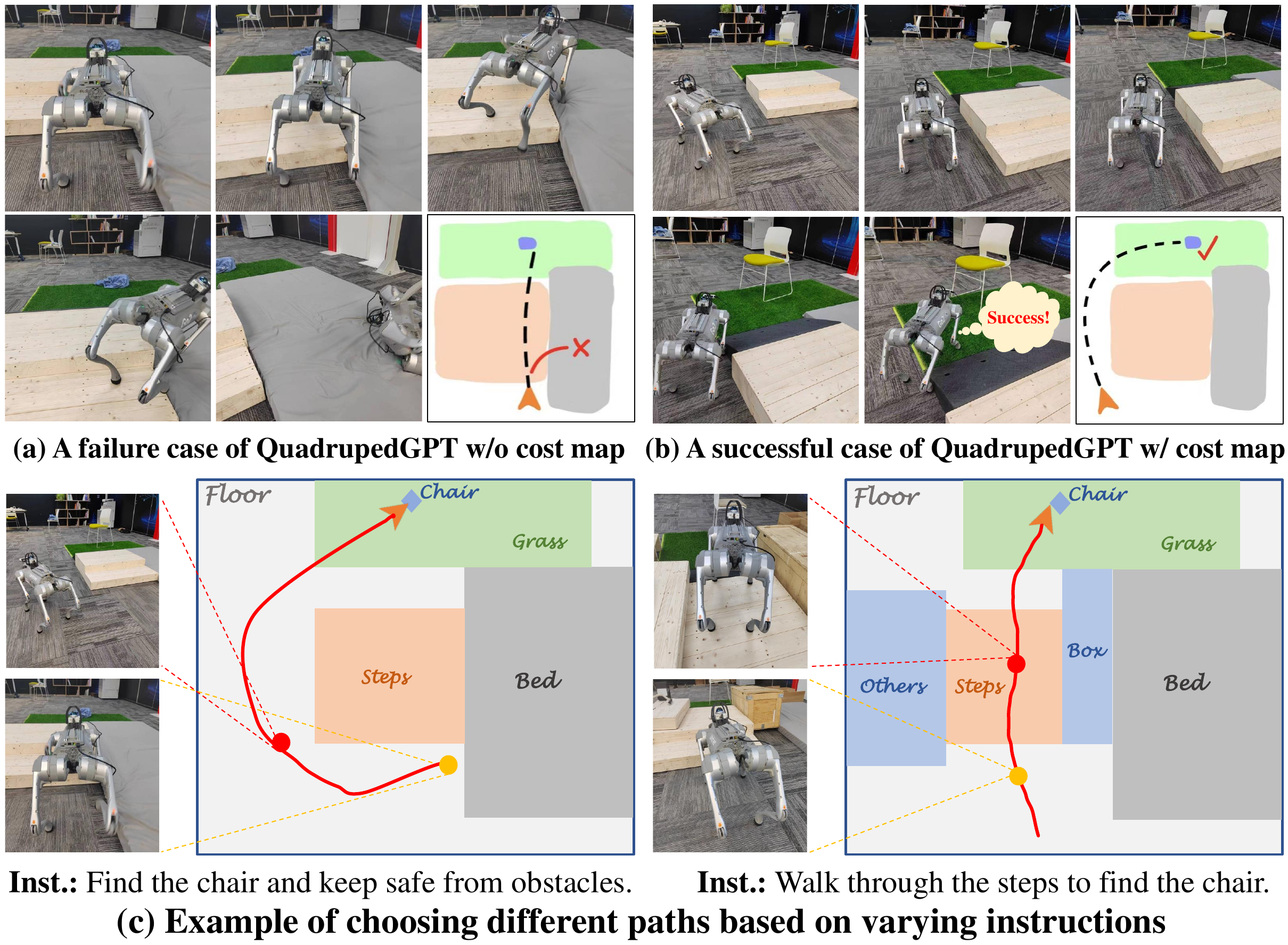}
  \caption{Cases of path planning with and without using open-vocabulary semantic cost map. Their semantic map examples are shown on the right side, in which different colors represent different objects or terrains. 
  }
  \label{fig:path_plan}
\end{figure}

\input{table/real_world_loco}

\subsection{Results of Path Planning}

\noindent\textit{\textbf{Cost map effectively helps the robot avoid challenging obstacles.} }
We conduct real-world experiments to evaluate the path planning performance of a quadruped agent targeting various objects under two distinct strategies.
The non-cost strategy disregards terrain costs, causing the agent to select the shortest path to the target.
This often leads to navigating through challenging terrains, increasing the risk of failure.
In contrast, our approach employs an LMM to assign costs to complex terrains, guiding the robot along smoother and safer paths.
As shown in Figure~\ref{fig:path_plan} (a) and (b), our cost map strategy outperforms the non-cost strategy. 
This improvement is largely due to the LMM's effectiveness in guiding the robot away from challenging obstacles, substantially reducing the likelihood of task failure. 
Without such a cost map, the agent struggles to adjust its path around obstacles, leading to failure, such as when navigating over a mattress.
Figure~\ref{fig:path_plan} (c) illustrates how varying instruction enables the LMM to assign different costs to terrains, resulting in varied paths.
For more details, please refer to the supplementary video.

\subsection{Results of High-level Reasoning}

\begin{figure}[t]
  \centering
  \includegraphics[width=\linewidth]{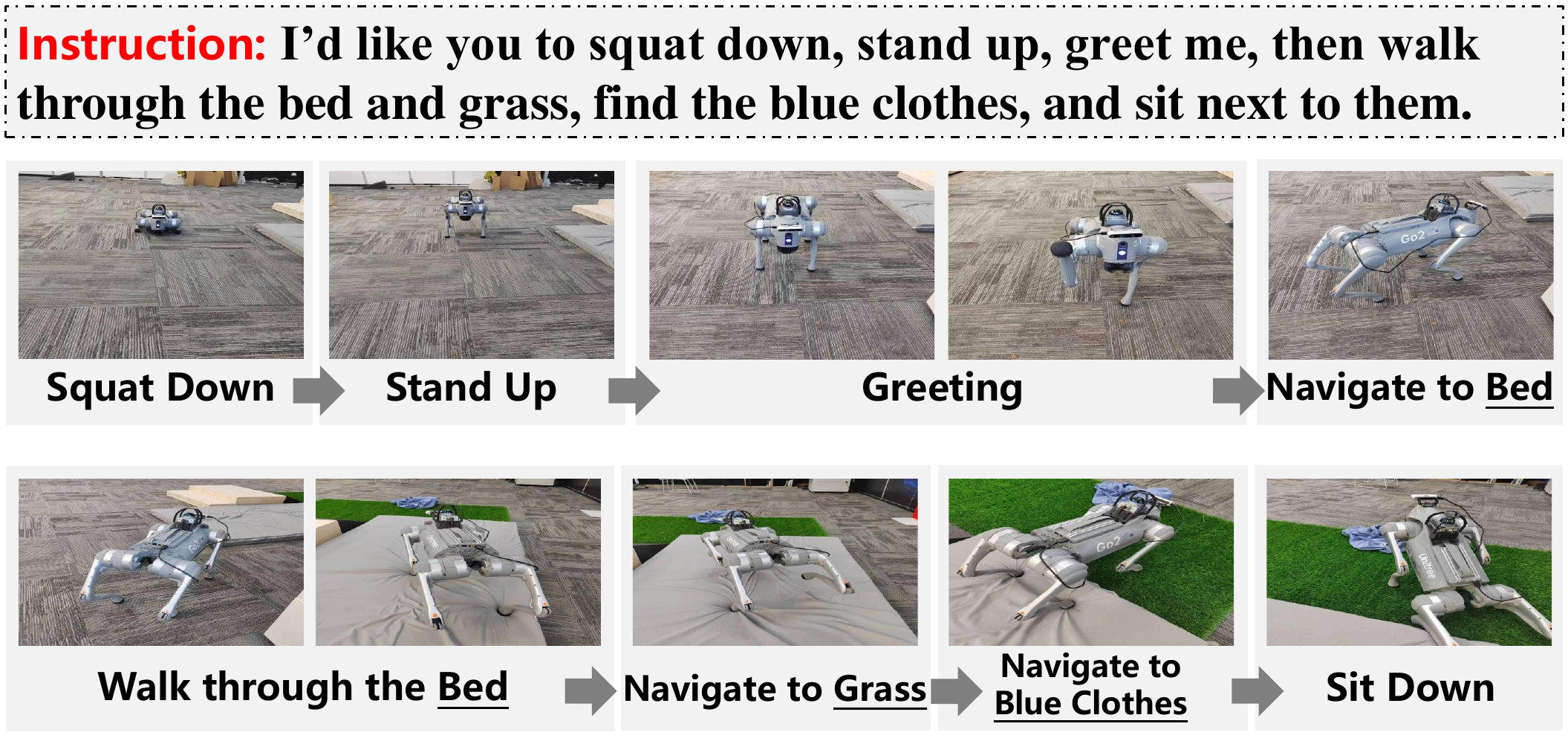}
  \caption{An example of the quadruped agent successfully completing the long-term objective based on a given human instruction.}
  \label{fig:high-level}
\end{figure}

To evaluate the high-level reasoning capabilities of our agent, we task the robot with completing a sequence of long-term goals (i.e., ``\textit{squat down, stand up, greet me, then walk through the bed and grass, find the blue clothes, and sit next to them}''). 
QuadrupedGPT needs to decompose the human instruction into subgoals (i.e., \textit{squat down}, \textit{find the blue clothes}) and search the right skill from the skill library to accomplish these subgoals in order. 
Figure \ref{fig:high-level} provides a qualitative example of the agent successfully completing the long-horizon task based on the given instruction.

\subsection{Discussion}

\noindent\textit{\textbf{Comparison with terrain-specific specialist policies. }} 
In locomotion experiments, we do not compare with single policies trained on specific terrains. This is due to the complexity involved in intricate reward engineering and well-structured learning curriculum design for such specialist policies. 
We find that challenging terrains introduce significant variance into the reinforcement learning process, leading to failures in training adaptable policies like~\cite{margolis2023walk}. 
Moreover, such specialist policies exhibit limited generalizability. 
Consequently, we adopt the adjustable policy combined with an automatic adaptation approach.

\noindent\textit{\textbf{Independent evaluation of modules.}} 
Our focus is on the overall performance of the agent, so we do not provide independent evaluations for each modules. For example, in locomotion experiments, we do not calculate the exact match metric for LMM selection. This is partly due to the difficulty in obtaining accurate ground truth choices and because multiple parameter sets may perform equally well in real-world environments, all leading to successful outcomes.

\noindent\textit{\textbf{Limitations. }}
Many factors, such as the accuracy of the depth camera and the segmentation model, still limit the current agent's performance. 
Our future work may focus on improving the precision and robustness of the perception module to improve the robot's navigation capabilities in complex real-world environments.

%% file: table/reward.tex
\begin{table}[h]
\centering
\caption{Details of rewards for locomotion adaption evaluation, where $C_{\rm foot}^{\rm cmd}(\theta, t)$ computes the desired contact state of each foot from the phase and timing variable~\cite{siekmann2021sim}.}
\scalebox{0.85}{
\begin{tabular}{lll}
\toprule
TERM & DEFINITION &  EQUATION   \\ 
\midrule

{\large $\displaystyle r_{v_{x,y}^{\rm cmd}}$}  & x, y velocity tracking & $\exp\{-|v_{x,y} - v_{x,y}^{\rm cmd}|^2/\sigma_{vxy}\}$  \\

{\large $\displaystyle r_{\omega_{z}^{\rm cmd}}$}  & yaw velocity tracking & $\exp\{-|\omega_z - \omega_z^{\rm cmd}|^2/\sigma_{\omega z}\}$ \\

{\large $\displaystyle r_{c_{f}^{\rm cmd}}$}  & swing phase tracking & $\sum_{\rm foot}[1-C_{\rm foot}^{\rm cmd}(\theta, t)]\exp\{-|{\rm f}^{\rm foot}|^2/\sigma_{cf}\}$  \\

{\large $\displaystyle r_{c_v^{\rm cmd}}$}  & stance phase tracking & $\sum_{\rm foot}[1-C_{\rm foot}^{\rm cmd}(\theta, t)]\exp\{-|{\rm v}^{\rm foot}_{xy}|^2/\sigma_{cv}\}$ \\

\bottomrule
\end{tabular}}
\label{tab:reward}
\end{table}

%% file: table/locomotion.tex
\begin{table*}[ht]
\centering
\caption{Results on the locomotion adaption benchmark in simulation, where ``Manual'', ``Auto'', ``LSS'' denote manual parameter tuning~\cite{margolis2023walk}, automatic locomotion adaptation by directly predicting parameter numbers, and by adopting the Location-Simulation-Selection strategy, respectively. 
All results are expressed as percentages of maximum episodic reward.}
\scalebox{0.85}{
\begin{tabular}{llllll}
\toprule
TERRAIN  &  METHOD  & {\Large $\displaystyle r_{v_{x,y}^{\rm cmd}}$}   & {\Large $\displaystyle r_{\omega_z^{\rm cmd}}$}    & {\Large $\displaystyle r_{c_f^{\rm cmd}}$}    & {\Large $\displaystyle r_{c_v^{\rm cmd}}$}  \\ 
\midrule

\multirow{5}{*}{\makecell[c]{Uphill Slope}}       
& Manual & 62.20 &63.83 &91.20 &95.35 \\
& Expert &70.08 &75.68 &95.44 &94.15 \\
\cline{2-6}
\noalign{\vskip 1.5mm}
& Auto        & 71.48 & 77.35 & \textbf{93.99} & 95.10 \\
& Auto+LSS & \textbf{71.83} (\textcolor{blue}{+9.63}) & \textbf{80.03} (\textcolor{blue}{+16.20}) & 93.95 (\textcolor{blue}{+2.75}) & \textbf{95.61} (\textcolor{blue}{+0.26}) \\
\midrule

\multirow{5}{*}{\makecell[c]{Downhill Slope}} 
& Manual &44.52 &52.53 &88.55 &90.58 \\
& Expert &69.91 &76.89 &93.07 &96.14 \\
\cline{2-6}
\noalign{\vskip 1.5mm}
& Auto                 & \textbf{73.07} & 75.29 & \textbf{93.42} & \textbf{96.24} \\
& Auto+LSS &71.76 \textcolor{blue}{(+27.24)}      &\textbf{76.59} \textcolor{blue}{(+24.06)}       &93.25 \textcolor{blue}{(+4.70)}       & 96.00 \textcolor{blue}{(+5.42)}      \\
\midrule

\multirow{5}{*}{\makecell[c]{Upward Stair}}   
& Manual &24.26 &40.99 &83.73 &91.91 \\
& Expert &29.44 &43.97 &90.69 &93.01 \\
\cline{2-6}
\noalign{\vskip 1.5mm}
& Auto                 & 25.98 & 39.01 & \textbf{87.77} & 91.72 \\ 
& Auto+LSS & \textbf{31.66} \textcolor{blue}{(+7.40)}       & \textbf{44.17} \textcolor{blue}{(+3.18)}      & 87.29 \textcolor{blue}{(+3.56)}      & \textbf{94.44} \textcolor{blue}{(+2.53)}      \\
\midrule

\multirow{5}{*}{\makecell[c]{Downward Stair}}   
& Manual &37.82 &43.95 &87.29 &91.92 \\
& Expert & 60.31 & 63.50 & 94.85  & 93.94 \\
\cline{2-6}
\noalign{\vskip 1.5mm}
& Auto                 & 55.62 & 60.41 & \textbf{93.03} & 91.63 \\ 
& Auto+LSS & \textbf{61.97} \textcolor{blue}{(+24.15)}      & \textbf{62.10} \textcolor{blue}{(+18.16)}      & 91.75 \textcolor{blue}{(+4.46)}      & \textbf{95.93} \textcolor{blue}{(+4.01)}      \\
\midrule
                                
\multirow{5}{*}{\makecell[c]{Uneven Ground}} 
& Manual &43.65 &44.45 &87.01 &92.74\\
& Expert &54.31 &56.07 &92.47 &94.22 \\
\cline{2-6}
\noalign{\vskip 1.5mm}
& Auto                 & 51.33 & 51.62 & \textbf{91.23} & 94.84 \\
& Auto+LSS & \textbf{53.56} (\textcolor{blue}{+9.91}) & \textbf{54.57} (\textcolor{blue}{+10.12}) & 90.65 (\textcolor{blue}{+3.64}) & \textbf{95.11} (\textcolor{blue}{+2.37}) \\

\bottomrule
\end{tabular}}
\label{tab:locomotion}
\end{table*}

%% file: table/real_world_loco.tex
\begin{table}[t]
\centering
\caption{Locomotion results in the real world. The success rate of 5 runs is reported. 
``FIXED'' denotes fixed average-parameter gait. ``LSS'' denotes automatic locomotion adaption. }
\scalebox{0.9}{
\begin{tabular}{m{1cm}<{\centering} m{3cm}<{\centering} m{1.7cm}<{\centering} m{1.7cm}<{\centering}}
\toprule
TERRAIN & ENVIRONMENT &  LSS  & FIXED   \\ 
\midrule

Uphill Slope 
& \includegraphics[width=0.7\linewidth]{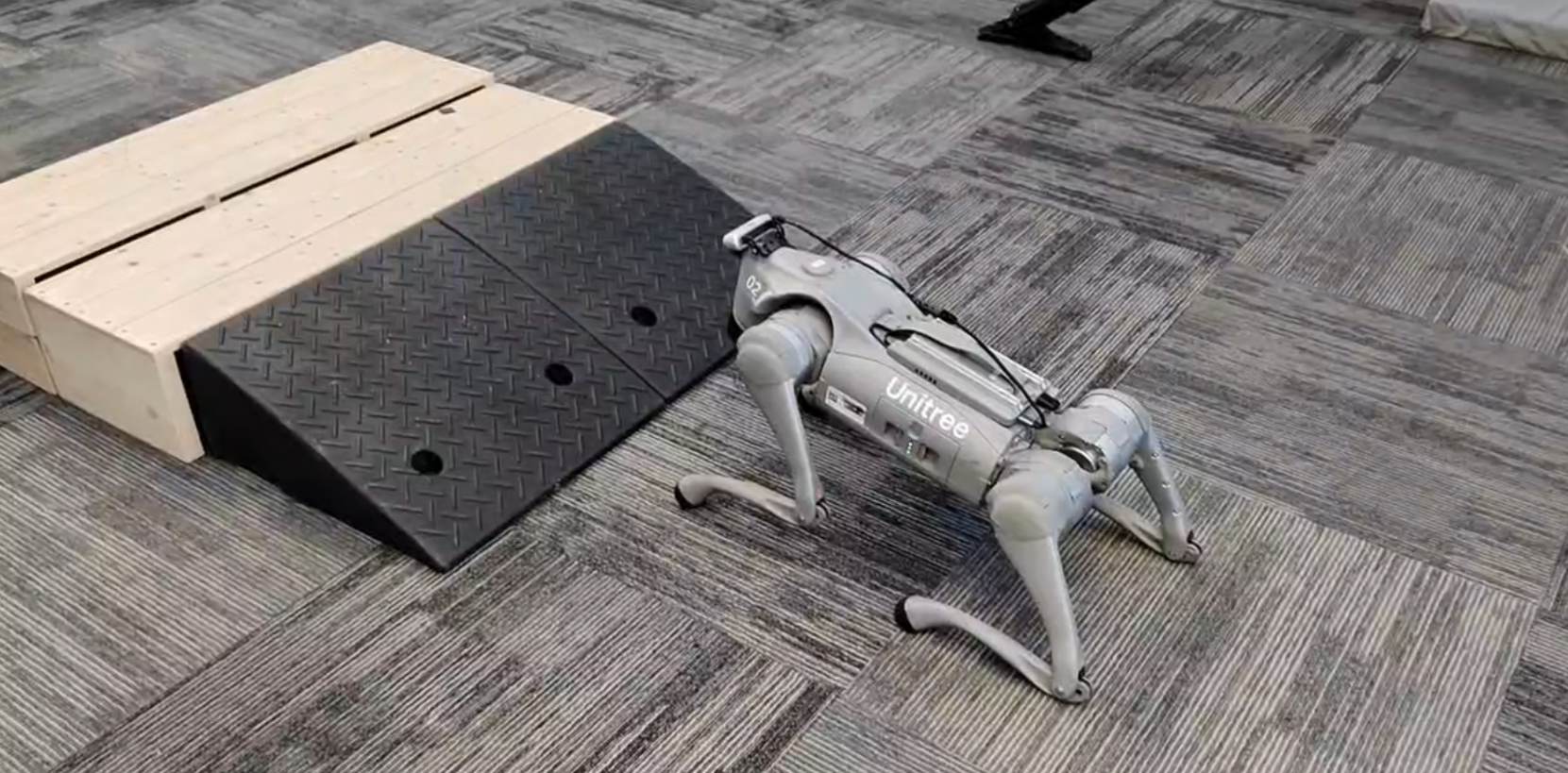}
& 4/5 & 2/5
\\
\midrule

Upward Stair  
& \includegraphics[width=0.7\linewidth]{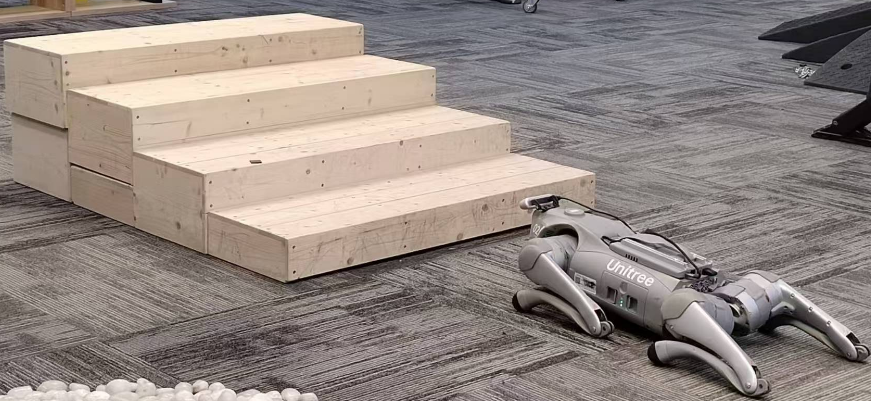}
& 4/5 & 0/5 
\\
\bottomrule
\end{tabular}}
\label{tab:real_world_loco}
\end{table}

%% file: Sections/5_conclusion.tex
\section{Conclusion}
\label{sec: conclusion}
QuadrupedGPT introduces a versatile quadruped agent that combines the agility of animals  with human-like cognitive abilities.
Leveraging the advanced comprehension abilities of LMMs, our agent autonomously navigates and interacts within an open-ended environment, effectively addressing critical challenges such as automatic locomotion adaption, efficient path planning, and complex reasoning.
Extensive evaluations demonstrate QuadrupedGPT's  proficiency in different benchmarks, showcasing its potential as a general-purpose quadruped agent suitable for practical applications in real-world scenarios.

%% file: Supplementary/Sections/all.tex
In this appendix, we first provide more details related to the benchmarks for locomotion adaption, including additional details of the specific terrains involved (Section~\ref{sec:terrain}), 
the prompt design (Section~\ref{sec:prompt}), 
the methodology for parameter sampling (Section~\ref{sec:param_sample}), and further analytical insights (Section~\ref{sec:add_analysis}).

Furthermore, we introduce the modular architecture of the path planning system implemented in QuadrupedGPT in Section~\ref{sec:path_overview}, whose architecture comprises several key components including the perception module (Section~\ref{sec:perception}), the semantic map representation (Section~\ref{sec:semantic_map}), the memory module for objects and terrains (Section~\ref{sec:obj_memory}), the trajectory planning module (Section~\ref{sec:traj_plan}),
and the design of prompt (Section~\ref{sec:prompt_for_cost}).

\subsection{Details of the Terrains}
\label{sec:terrain}

\input{Supplementary/Tables/terrains}

In our study focusing on the locomotion adaption benchmark, we select the Isaac Gym environment~\cite{makoviychuk2021isaac} to simulate various challenging terrains, which is well-suited for testing the dynamic responses of robots to different ground conditions.
As shown in Table~\ref{tab: terrain}, we provide a comprehensive parameter setup for each type of terrain included in this benchmark.
Specifically, for terrains where the quadruped robot needs to adjust its stability by lowering its center of gravity (e.g., downhill slopes and downside stairs), we have designed a slightly larger central platform area.
This modification aims to offer longer preparation time for the robot as it traverses these challenging terrains, allowing for precise adjustments to the robot's behavior in response to varying environmental conditions.

\subsection{Details of Prompt Design for LSS}
\label{sec:prompt}
In this section, we introduce how to curate the prompts for the automatic parameter generation process ``Auto'' and ``Auto+LSS''.
Table~\ref{tab: para_range} provides detailed information on the specific ranges for parameter adjustments to curate the prompts.

\begin{figure*}[htb]
  \centering
  \includegraphics[width=\linewidth]{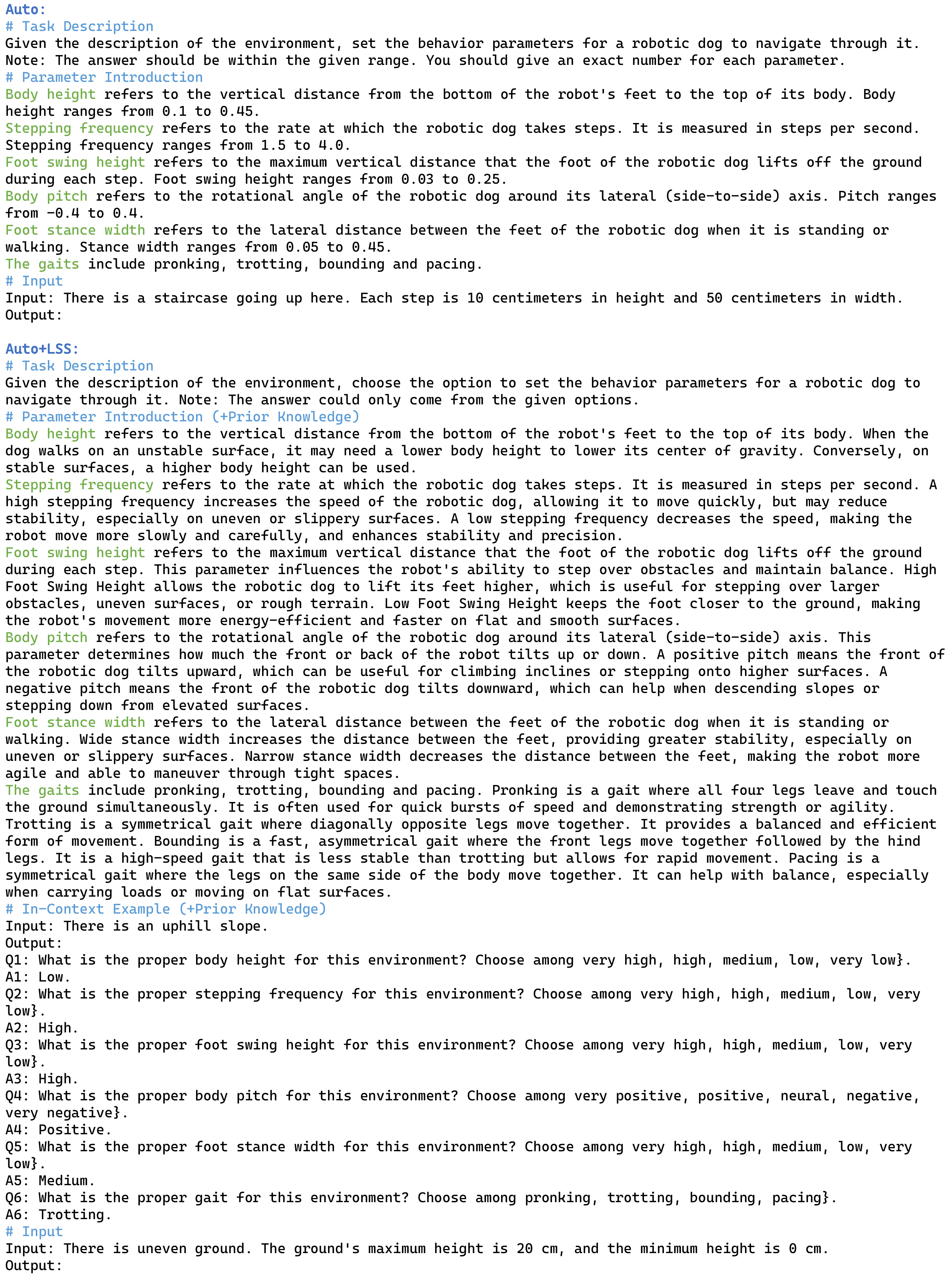} 
  \caption{Prompt examples of LSS.}
  \label{fig:loco_prompt} 
\end{figure*}

As introduced in the main paper, ``Auto'' describes the method where the LMM directly generates numerical parameters essential for locomotion skills. 
In contrast, ``Auto+LSS'' employs a more sophisticated approach known as the Location-Simulation-Selection (LSS) strategy.
During the ``Location'' phase of ``Auto+LSS'', the LMM is first tasked with determining appropriate ranges for the parameters as detailed in Table \ref{tab: para_range}.
Following this, we utilize GPT-4o~\cite{gpt4o} to generate three candidate sets of parameters.
From these candidates, we either compute the average of the results (``Auto'') or select the range that receives the majority of votes (``Auto+LSS'') for implementation in our experiments.
This strategy ensures that the generated parameters are applicable for diverse simulation environments.
To illustrate our prompt design according to the above process, we provide prompt examples for both ``Auto'' and ``Auto+LSS'', respectively in Figure \ref{fig:loco_prompt}.

\input{Supplementary/Tables/para_range}

\subsection{Parameter Sampling and Selection}
\label{sec:param_sample}
In our experimental setup, we employ uniform sampling to determine values for all continuous adjustable parameters within the ranges specified by the LMM.
Specifically, we set sampling intervals for key locomotive parameters as: body height at 0.05, stepping frequency at 0.2, foot swing height at 0.02, body pitch at 0.08, and foot stance width at 0.05.
Once these parameters are sampled, they are permuted and evaluated in our simulation environment to assess their impact on the robot's performance.
The effectiveness of each parameter set is evaluated based on the performance metric $r_{v_{x,y}^{\rm cmd}}$, which is calculated as follows:

\begin{equation}
r_{v_{x,y}^{\rm cmd}} = \exp\left(\frac{-|v_{xy}-v_{xy}^{\rm cmd}|^2}{\sigma_{vxy}}\right)
\end{equation}

This formula measures the deviation of the robot's actual velocity ($v_{xy}$) from its commanded velocity ($v_{xy}^{\rm cmd}$), normalized by a variance term ($\sigma_{vxy}$).
The parameter set yielding the highest $r_{v_{x,y}^{\rm cmd}}$ value is selected for implementation, as it demonstrates the best speed performance under current terrain conditions.

\subsection{Further Analysis}
\label{sec:add_analysis}
\input{Supplementary/Tables/locomotion_adaption}

\textbf{Does prior knowledge on the basis of Auto help? }
Based on the results presented in Table~\ref{tab: locomotion_result}, the inclusion of prior knowledge in the form of detailed parameter explanations and contextual examples does not enhance, but rather diminishes the performance of the ``Auto'' strategy.
This unexpected decline may be attributed to the LMM's limitations in effectively processing numerical values.
As the input becomes more complex and verbose, the LMM struggles to capture the extended critical information.
This observation highlights the superior efficacy of our Location-Simulation-Selection (LSS) strategy in managing continuous inputs.

\textbf{How does manual tuning affect the results? } 
Manual tuning introduces a significant variability in locomotion outcomes, which often pushes parameter values toward more extreme limits compared to automatic strategies.
For instance, in our experiment involving ten participants tasked with manually setting parameters on the Downside Stair terrain, a notable variability in their selection is observed.
The average body pitch determined by the participants is $-0.3$, with four of them giving the lowest possible value of $-0.4$.
Such results contrast sharply with the range selected by the LMM, which is only between $[-0.24, -0.08]$, a comparatively moderate range.
The tendency of manual tuning to skew toward extreme values suggests its potential limitations, particularly for those human pilots who are less experienced.
Given that the robotic control policies are highly sensitive to these adjustable parameters, the experimental results shown in Table~\ref{tab: locomotion_result} indicate the challenge and risks associated with manual tuning in achieving optimal performance.

\subsection{Overview of Path Planning Module}
\label{sec:path_overview}
Our path planning module is developed using a modular architecture based on the GOAT framework~\cite{chang2023goat}.
Unlike the original implementation, our QuadrupedGPT extends its functionality to not only detect specific object instances but also to categorize different types of terrain.
These elements are identified and localized within a top-down semantic map of the scene, which is then stored in a dedicated semantic memory.
The path planning process begins when a goal is specified.
The global policy first searches the semantic memory to locate the goal.
If the goal is not found within the stored memory, the global policy shifts its focus to exploration, setting an interim exploration goal to gather more information about the environment. 
Subsequently, a local policy takes over to compute the atomic actions required to navigate towards the long-term goal, optimizing the path based on the dynamically updated semantic map and current situational awareness.

\subsection{Perception Module}
\label{sec:perception}
For sensory input and environmental mapping, we employ the RealSense D435i camera, which captures RGB images ($I_t$) and depth images ($D_t$) at each timestep.
In addition, onboard sensors provide 2D positional data and RPY (roll, pitch, yaw) orientation. 
The RGB image $I_t$ is initially processed using the LMM to identify object and terrain categories.
These categories are then utilized as semantic labels for the dense segmentator, where we adopt SEEM~\cite{zou2023segment} in this paper. 
Following segmentation, the labels are projected onto a point cloud generated from the depth image $D_t$. 
This point cloud is then binned into a 3D semantic voxel map, which segments the environment spatially into bins that incorporate semantic data with geometric structure. 
To synthesize this information into a practical format for navigation and planning, we perform a vertical summation over the voxel map's height, resulting in a comprehensive 2D instance map ($m_t$).
This map provides a detailed representation of the terrain and object distribution at ground level, which is crucial for path planning and obstacle avoidance in dynamic environments.

\subsection{Semantic Map Representation}
\label{sec:semantic_map}
The semantic map, denoted as $m_t$ for each timestep $t$, serves as a spatial representation of the environment and is crucial for tracking the coverage of various object and terrain types. 
The map is structured as a $K \times M \times M$ integer matrix, where $M \times M$ represents the map's area and $K$ is the total number of map channels.
Specifically, each cell within this grid corresponds to a real-world area of 25 square centimeters (5 cm $\times$ 5 cm).
The channels of the map are defined by $K = C + 3$, where $C$ is the number of distinct semantic categories identified in the environment.
The additional three channels are designed for logging explored areas, as well as the agent's current and historical positions, respectively.
In this map, a non-zero entry in a particular channel indicates that the cell contains either an object or terrain region corresponding to that semantic category, or that the cell has been explored.
Conversely, a zero value indicates that the cell has not been classified as belonging to that semantic category or has not yet been explored.
The first $C$ channels of the map store the unique instance IDs of the objects as they are projected onto the map.
At the onset of each episode, the map is initialized to all zeros, with the robot positioned centrally, facing east.
This initialization sets the stage for the agent to systematically explore and interact with its environment, progressively filling the map with valuable semantic and positional data.

\subsection{Memory Module for Objects and Terrains}
\label{sec:obj_memory}
Our memory module clusters object and terrain detection results over time, grouping them into distinct instances based on their location on the map and their category.
Each instance, denoted by $i$, is cataloged as a set of map cells $C_i$, a collection of object views represented as bounding boxes with contextual details $M_i$, and an integer indicating the semantic category $S_i$.
When a new RGB image is processed, objects and terrains within it are detected.
For each detected element $d$, we define the bounding box $I_d$ around the detection, assign a semantic class $S_d$, and map the corresponding points $C_d$ based on projected depth information.
To enhance the matching process between new detection and existing memory entries, each set of points $C_d$ on the map is dilated by $p$ units, creating an expanded set of points $D_d$.
This dilation allows for a broader search area when matching new detection with previously recorded instances.

In addition, pairwise matching checks are conducted between each new detection and existing instances in the memory. 
A detection $d$ is considered to match an instance $i$ if they share the same semantic category and there is any overlap between the dilated detection area and the instance's location on the map, i.e., $S_d = S_i$ and $D_d \cap C_i \neq \emptyset$.
Upon finding a match, we update the existing instance by merging the new points and image into it:
\begin{equation}
C_i \leftarrow C_i \cup C_d, \ M_i \leftarrow M_i \cup {I_d}
\end{equation}

If no existing instance matches the detection, a new instance is created using $C_d$ and $I_d$.
This systematic aggregation of unique instances over time facilitates efficient matching of new goals to all relevant images and categories previously cataloged. 
Thus, the dynamic memory structure can enhance the robustness and accuracy of our path planning by maintaining a comprehensive and evolving understanding of the environment.

\begin{figure*}[htb]
  \centering
  \includegraphics[width=\linewidth]{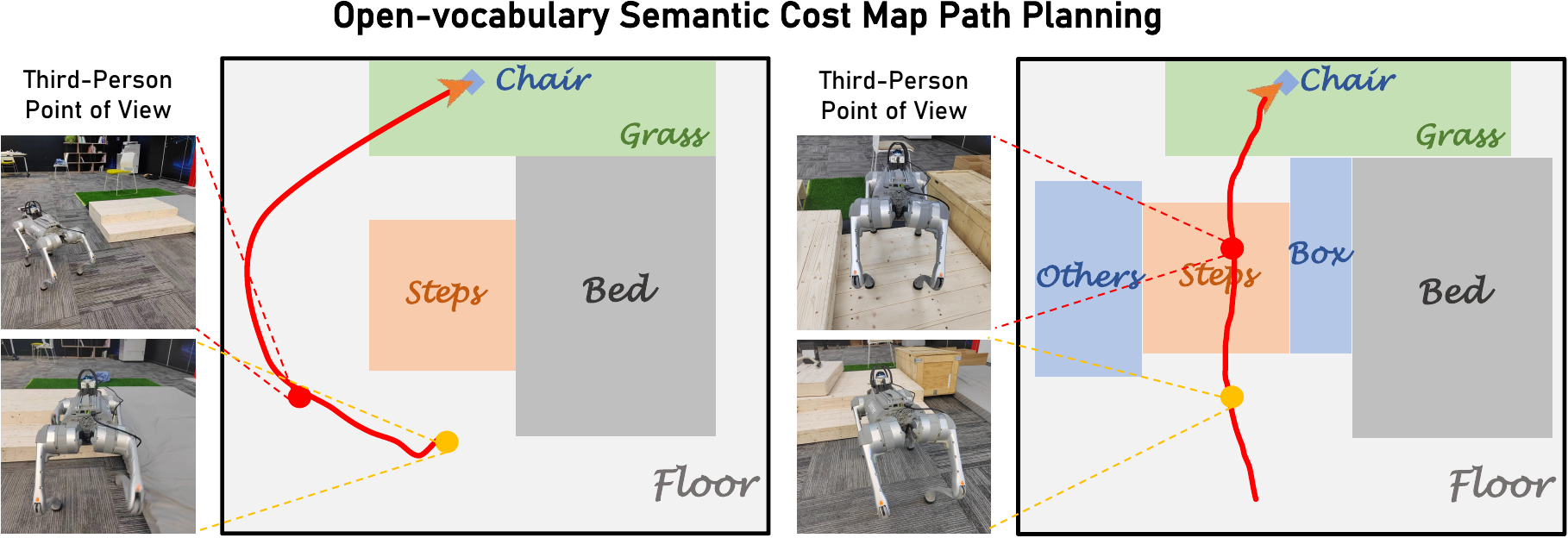} 
  \caption{The figure illustrates a schematic diagram of the semantic cost map and planned path for a quadruped robot during path planning. The left side showcases the selection of the safest path, while the right side demonstrates choosing the shortest path while avoiding obstacles in a more complex environment.}
  \label{fig:semantic map} 
\end{figure*}

\subsection{Trajectory Planning Module}
\label{sec:traj_plan}
When a new goal is specified to the agent, the global policy ($\pi_G$) initiates by searching the instance memory to check if the goal has already been encountered and recorded.
The matching process between the specified goal and memory instances is adapted according to the modality of the goal input. 
If a matching instance is found, its recorded location on the top-down map is set as a long-term navigation goal.
If no such instance is found, $\pi_G$ shifts its strategy to generate an exploration goal, employing a frontier-based approach that targets the nearest unexplored region.

Once the global policy identifies a long-term goal, the local policy ($\pi_L$) takes over.
It begins by assigning differential costs given by LMM to various semantic categories represented on the semantic map, effectively constructing a nuanced cost map.
This map forms the basis for path planning, where the Fast Marching Method is applied to calculate an optimal route to the goal.
On the Go2 robot, the process is achieved by calculating the positional offset from the robot's current location to the target point and employing its built-in motion control system to navigate to the target.

Figure~\ref{fig:semantic map} illustrates the path planning scheme of the local policy $\pi_L$ after a long-term goal is determined.
The left side of the figure shows a cost map that prioritizes the safest path.
The robot chooses to bypass difficult terrains, opting for the safest "floor" to reach the target.
The right side demonstrates a more complex environment with a higher density of objects, where the cost of descending stairs is lowered.
This encourages the robot to choose the fastest stair path while also avoiding other obstacles and promptly switching gaits to reach the designated goal.

\subsection{Details of Prompt Design for Cost Map Construction}
\label{sec:prompt_for_cost}
\begin{figure*}[htb]
  \centering
  \includegraphics[width=\linewidth]{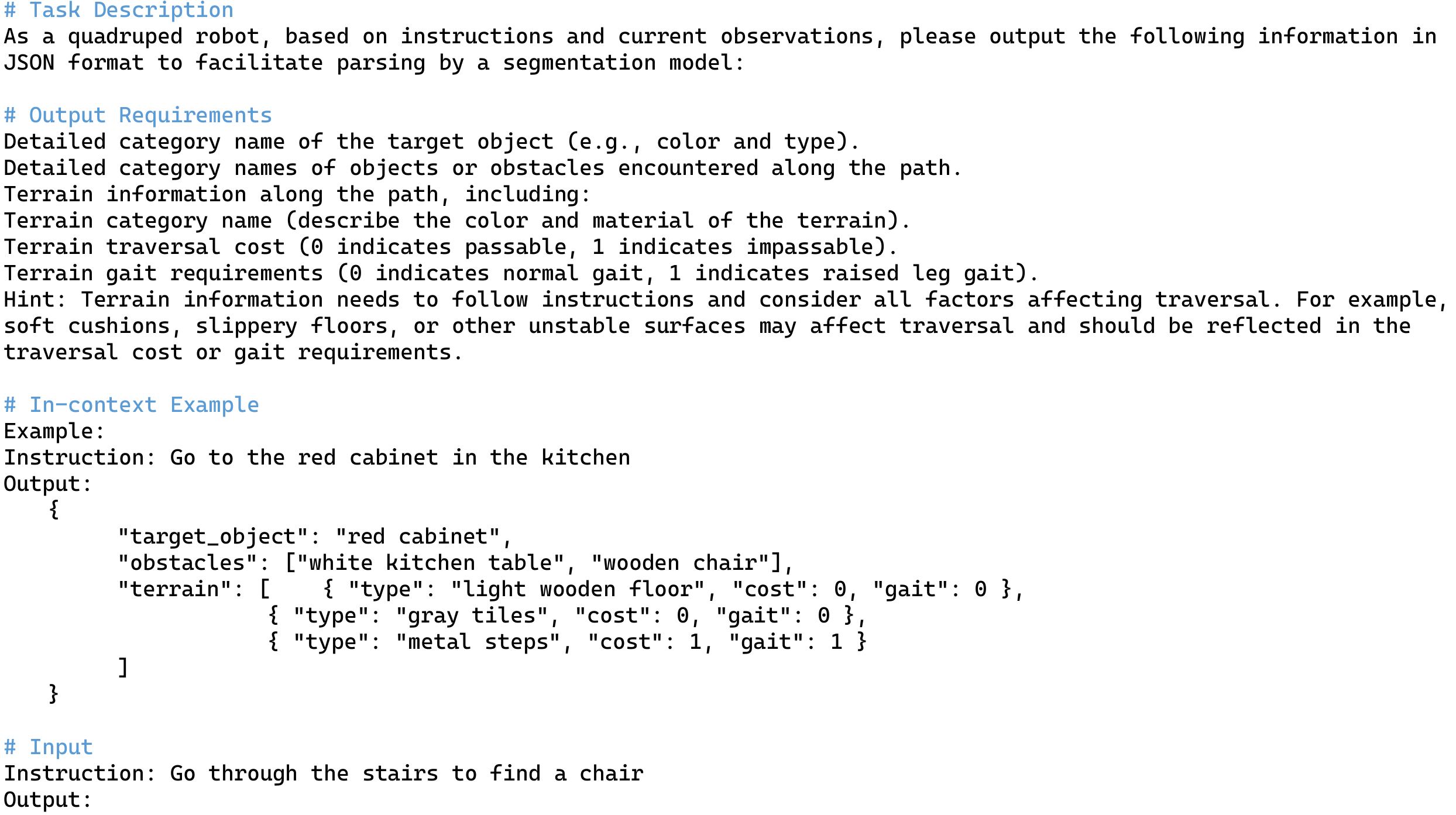} 
  \caption{Prompt examples for cost map construction.}
  \label{fig:cost_prompt} 
\end{figure*}

In this section, we introduce how to curate the prompts for cost map construction. We provide prompt example in Figure ~\ref{fig:cost_prompt}.

%% file: Supplementary/Tables/terrains.tex
\begin{table*}[h]
\centering
\caption{The parameter setup for each type of the terrains included in the locomotion adaption benchmark, where the length is measured in meters.}
\label{tab: terrain}
\setlength{\tabcolsep}{12pt}
\scalebox{1}{
\begin{tabular}{ll}
\toprule
TERRAIN       & PARAMETER                                               \\
\midrule
Uphill Slope   & ${\rm slope}=-0.15,\: {\rm platform\: size}=0.6$ \\[3pt]
Downhill Slope & ${\rm slope}=0.4,\: {\rm platform}\: {\rm size=0.8}$  \\[3pt]
Upside Stair   & ${\rm step}\: {\rm width}=0.5, {\rm step}\: {\rm height}=-0.1, {\rm platform}\: {\rm size}=0.8$ \\[3pt]
Downside Stair & ${\rm step}\: {\rm width}=0.5, {\rm step}\: {\rm height}=0.1, {\rm platform}\: {\rm size}=1.0$   \\[3pt]
Uneven Ground  & ${\rm min}\: {\rm height}=0, {\rm max}\: {\rm height}=0.2$ \\
\bottomrule
\end{tabular}}
\end{table*}

%% file: Supplementary/Tables/para_range.tex
\begin{table*}[h]
\centering
\caption{Specific ranges of different adjustable parameters to set up the prompt.}
\label{tab: para_range}
\begin{tabular}{clc|clc}
\toprule
  & \multicolumn{2}{c|}{LOCATION RANGE} & & \multicolumn{2}{c}{LOCATION RANGE} \\
\midrule
\multirow{5}{*}{Body Height (m)}         & Very High      & 0.4$\sim$0.45 & \multirow{5}{*}{\makecell{Stepping \\Frequency (Hz)}} & Very High  & 3.5$\sim$4    \\[1.5pt]
& High   & 0.3$\sim$0.4  && High       & 3$\sim$3.5     \\[1.5pt]
& Medium & 0.2$\sim$0.3  && Medium    & 2.5$\sim$3    \\[1.5pt]
& Low   & 0.15$\sim$0.2   && Low      & 2$\sim$2.5   \\[1.5pt]
& Very Low & 0.1$\sim$0.15   && Very Low  & 1.5$\sim$2   \\
\midrule

\multirow{5}{*}{Body Pitch (rad)} & Very Positive  & 0.24$\sim$0.4  & \multirow{5}{*}{\makecell{Foot Swing \\Height (m)}} & Very High & 0.21$\sim$0.25    \\[1.5pt]
& Positive  &   0.08$\sim$0.24  && High   & 0.16$\sim$0.21   \\[1.5pt]
& Neural    &   -0.08$\sim$0.08  && Medium   & 0.11$\sim$0.16     \\[1.5pt]
& Negative   &    -0.24$\sim$-0.08 && Low    & 0.07$\sim$0.11   \\[1.5pt]
& Very Negative  &     -0.4$\sim$-0.24  && Very Low  & 0.03$\sim$0.07 \\
\midrule

\multirow{5}{*}{Gait Type ($\theta$)}  & Pronking & $[0, 0, 0]$ & \multirow{5}{*}{\makecell{Foot Stance \\Width (m)}}   & Very High    & 0.37$\sim$0.45    \\[1.5pt]
& Trotting       & $[0.5, 0, 0]$ && High        & 0.29$\sim$0.37    \\[1.5pt]
& Bounding       & $[0, 0.5, 0]$ && Medium     & 0.21$\sim$0.29    \\[1.5pt]
& Pacing         & $[0, 0, 0.5]$ && Low       & 0.13$\sim$0.21  \\[1.5pt]
&                &             && Very Low       & 0.05$\sim$0.13  \\
\bottomrule
\end{tabular}
\end{table*}

%% file: Supplementary/Tables/locomotion_adaption.tex
\begin{table*}[htb]
\centering
\caption{Additional results of the locomotion adaption benchmark. The terms ``Manual'', ``Auto'', ``LSS'' represents different tuning strategies: manual parameter tuning as introduced in~\cite{margolis2023walk}, automatic locomotion adaptation through direct numeric prediction, and the implementation of the Location-Simulation-Selection strategy, respectively. 
All outcomes are presented as percentages of maximum episodic reward achieved.}
\label{tab: locomotion_result}

\scalebox{0.95}{
\begin{tabular}{llllll}
\toprule
TERRAIN  &  METHOD  & {\Large $\displaystyle r_{v_{x,y}^{\rm cmd}}$}   & {\Large $\displaystyle r_{\omega_z^{\rm cmd}}$}    & {\Large $\displaystyle r_{c_f^{\rm cmd}}$}    & {\Large $\displaystyle r_{c_v^{\rm cmd}}$}  \\ 
\midrule

\multirow{5}{*}{\makecell[c]{Uphill Slope}}       
& Manual & 62.20 &63.83 &91.20 &95.35 \\[1.5pt]
& Expert &70.08 &75.68 &95.44 &94.15 \\[1.5pt]
\cline{2-6}
\noalign{\vskip 1.5mm}
& Auto        & 71.48 & 77.35 & \textbf{93.99} & 95.10 \\[1.5pt]
& Auto+prior  & 66.73 & 72.74 & 93.53 & 94.27 \\[1.5pt]
& Auto+LSS & \textbf{71.83} (\textcolor{blue}{+9.63}) & \textbf{80.03} (\textcolor{blue}{+16.20}) & 93.95 (\textcolor{blue}{+2.75}) & \textbf{95.61} (\textcolor{blue}{+0.26}) \\[1.5pt]
\midrule

\multirow{5}{*}{\makecell[c]{Downhill Slope}} 
& Manual &44.52 &52.53 &88.55 &90.58 \\[1.5pt]
& Expert &69.91 &76.89 &93.07 &96.14 \\[1.5pt]
\cline{2-6}
\noalign{\vskip 1.5mm}
& Auto                 & \textbf{73.07} & 75.29 & \textbf{93.42} & \textbf{96.24} \\[1.5pt]
& Auto+prior           & 64.84 & 67.31 & 92.40  & 95.61 \\[1.5pt]
& Auto+LSS &71.76 \textcolor{blue}{(+27.24)}      &\textbf{76.59} \textcolor{blue}{(+24.06)}       &93.25 \textcolor{blue}{(+4.70)}       & 96.00 \textcolor{blue}{(+5.42)}      \\[1.5pt]
\midrule

\multirow{5}{*}{\makecell[c]{Upside Stair}}   
& Manual &24.26 &40.99 &83.73 &91.91 \\[1.5pt]
& Expert &29.44 &43.97 &90.69 &93.01 \\[1.5pt]
\cline{2-6}
\noalign{\vskip 1.5mm}
& Auto                 & 25.98 & 39.01 & \textbf{87.77} & 91.72 \\[1.5pt] 
& Auto+prior           & 20.63 & 36.99 & 83.95 & \textbf{95.00}    \\[1.5pt]
& Auto+LSS & \textbf{31.66} \textcolor{blue}{(+7.40)}       & \textbf{44.17} \textcolor{blue}{(+3.18)}      & 87.29 \textcolor{blue}{(+3.56)}      & \textbf{94.44} \textcolor{blue}{(+2.53)}      \\[1.5pt]
\midrule

\multirow{5}{*}{\makecell[c]{Downside Stair}}   
& Manual &37.82 &43.95 &87.29 &91.92 \\[1.5pt]
& Expert & 60.31 & 63.50 & 94.85  & 93.94 \\[1.5pt]
\cline{2-6}
\noalign{\vskip 1.5mm}
& Auto                 & 55.62 & 60.41 & \textbf{93.03} & 91.63 \\ [1.5pt]
& Auto+prior           & 53.56 & 57.02 & 91.71 & 93.37 \\[1.5pt]
& Auto+LSS & \textbf{61.97} \textcolor{blue}{(+24.15)}      & \textbf{62.10} \textcolor{blue}{(+18.16)}      & 91.75 \textcolor{blue}{(+4.46)}      & \textbf{95.93} \textcolor{blue}{(+4.01)}      \\[1.5pt]
\midrule
                                
\multirow{5}{*}{\makecell[c]{Uneven Ground}} 
& Manual &43.65 &44.45 &87.01 &92.74\\[1.5pt]
& Expert &54.31 &56.07 &92.47 &94.22 \\[1.5pt]
\cline{2-6}
\noalign{\vskip 1.5mm}
& Auto                 & 51.33 & 51.62 & \textbf{91.23} & 94.84 \\[1.5pt]
& Auto+prior           & 45.26 & 53.50  & 90.21 & 92.63 \\[1.5pt]
& Auto+LSS & \textbf{53.56} (\textcolor{blue}{+9.91}) & \textbf{54.57} (\textcolor{blue}{+10.12}) & 90.65 (\textcolor{blue}{+3.64}) & \textbf{95.11} (\textcolor{blue}{+2.37}) \\[1.5pt]

\bottomrule
\end{tabular}}
\end{table*}